\documentclass[conference]{IEEEtran}
\IEEEoverridecommandlockouts
\usepackage{cite}
\usepackage{amsmath,amssymb,amsfonts}
\usepackage{algorithmic}
\usepackage{graphicx}
\usepackage{subfigure}
\usepackage{textcomp}
\usepackage{xcolor}
\usepackage{algorithm}
\usepackage{theorem}
\usepackage{algorithmic}
\usepackage{amsmath}
\usepackage{multirow}
\usepackage{palatino}
\usepackage{ccaption}
\usepackage{tikz}
\usepackage{amsfonts,amssymb}
\usepackage[T1]{fontenc} 

\newtheorem{theorem}{Theorem}
\newtheorem{definition}{Definition}

\newtheorem{lemma}{Lemma}

\def\BibTeX{{\rm B\kern-.05em{\sc i\kern-.025em b}\kern-.08em
    T\kern-.1667em\lower.7ex\hbox{E}\kern-.125emX}}
\DeclareRobustCommand*{\IEEEauthorrefmark}[1]{%
    \raisebox{0pt}[0pt][0pt]{\textsuperscript{\footnotesize\ensuremath{#1}}}}    
\begin{document}

\title{Adap DP-FL: Differentially Private
Federated Learning with Adaptive Noise \\
\thanks{* The corresponding author is Zhili Chen}
\thanks{This work is sponsored by the Natural Science Foundation of Shanghai (Grant NO. 22ZR1419100) and the National Natural Science Foundation of China (Grant NO. 62132005).}
}



\author{
\IEEEauthorblockN{
Jie Fu\IEEEauthorrefmark{},
Zhili Chen\IEEEauthorrefmark{*},
Xiao Han\IEEEauthorrefmark{}}
\IEEEauthorblockA{71215902083@stu.ecnu.edu.cn,  zhlchen@sei.ecnu.edu.cn, 51215902135@stu.ecnu.edu.cn}
\IEEEauthorblockA{\IEEEauthorrefmark{}Shanghai Key Laboratory of Trustworthy Computing, East China Normal University, Shanghai, China}
}
\maketitle

\begin{abstract}
Federated learning seeks to address the issue of isolated data islands by making clients disclose only their local training models. However, it was demonstrated that private information could still be inferred by analyzing local model parameters, such as deep neural network model weights. Recently, differential privacy has been applied to federated learning to protect data privacy, but the noise added may degrade the learning performance much. Typically, in previous work, training parameters were clipped equally and noises were added uniformly. The heterogeneity and convergence of training parameters were simply not considered. In this paper, we propose a differentially private scheme for federated learning with adaptive noise (Adap DP-FL). Specifically, due to the gradient heterogeneity, we conduct adaptive gradient clipping for different clients and different rounds; due to the gradient convergence, we add decreasing noises accordingly. Extensive experiments on real-world datasets demonstrate that our Adap DP-FL outperforms previous methods significantly.
\end{abstract}


\section{Introduction}
Nowadays, machine learning is widely applied in industry, and consumes a large amount of data. As a result, people concern more and more about the security and privacy of their own data. In 2007, to protect data privacy in machine learning, Google Inc. introduced the concept of federated learning (FL) for the first time\cite{mcmahan2017communication}. In such federated learning, multiple data holders (such as cell phones and IoT devices) collaborate to train models without sharing raw data, exchanging only training parameters. However, exposing training parameters instead of raw data is insufficient for privacy protection. It has been shown that, training parameters, such as gradient values, can be used to restore a portion of the original data\cite{zhu2019deep} or to infer whether the content of the mastered records is from a specific participant\cite{song2017machine}. Melis et al \cite{LucaMelis2022ExploitingUF} also demonstrated that shared models in FL can leak unexpected information about participants' training data. Therefore, extra measures should be taken to protect data privacy in federated learning.

\begin{figure}[htbp]
	\centering
	\subfigure[ gradient $l_2$-norm in Mnist dataset]{
	\includegraphics[width=3.8cm,height=2.5cm]{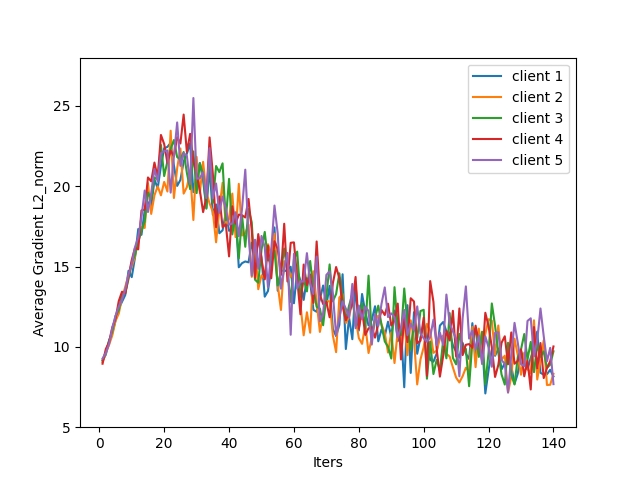}
	}
	\quad
	\subfigure[gradient $l_2$-norm in FashionMnist dataset]{
	\includegraphics[width=3.8cm,height=2.5cm]{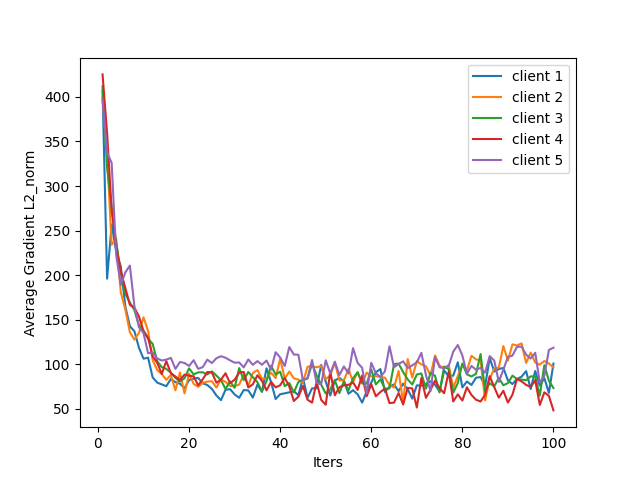}
	}
	\caption{The gradient $l_2$-norm vary with different clients and different rounds in federated learning}\label{fig:gradient-norm}
\end{figure}

In the last decade, differential privacy (DP) \cite{dwork2014algorithmic} has become the standard of privacy preservation, and it is extensively employed to protect privacy in machine learning, such as deep learning systems \cite{MartnAbadi2016DeepLW}. Recently, DP has also been applied in federated learning to achieve data privacy \cite{TheodorosSalonidis2019DifferentialPF,QingLiao2020DPFLAN,QinqingZheng2021FederatedP,KangWei2021UserLevelPF,HBrendanMcMahan2017LearningDP,RobinCGeyer2017DifferentiallyPF,StaceyTruex2020LDPFedFL,KangWei2019FederatedLW}. In existing differential private methods for federated learning, uniform noise was normally added everywhere and whenever. However, our observation is that the magnitudes of training parameters are heterogeneous in different places, and vary as training time goes. For example, Fig.~\ref{fig:gradient-norm} shows that in federated learning the gradient $l_2$-norm may vary with different client ends and training rounds, and finally converge to a small value. As a result, uniform noise may damage the training accuracy significantly due to gradient heterogeneity.

Motivated by the above observation, we study differentially private federated learning with two improvements as follows. 
1) gradient are clipped with adaptive clipping thresholds due to the heterogeneity of gradient magnitudes in term of client ends and training rounds. The gradient clipping is a subtlety for machine learning with differential privacy. Setting a too low clipping threshold may result in high bias due to the magnitude information loss. Setting a too high one, on the other hand, can result in severe noise addition. Both cases would damage the model utility greatly. Thus, the heterogeneity of gradient requires an adaptive gradient clipping to avoid the above two cases.
2) Noise scale is gradually decreased (resp. privacy budget is gradually increased) due to the convergence of gradient in term of training rounds. The intuition is that at the beginning of training, the model is far from optimization, and the gradient magnitudes are normally large, so greater noise is allowed. As the training proceeds, the model is approaching optimization and the gradient magnitudes are converging, smaller noise is required. 

Based on the above analysis, in this paper we propose a differentially private scheme for federated learning with adaptive
noise. We perform adaptive gradient clipping for different clients and different rounds, and decrease the noise scale adaptively as the training proceeds. Specifically, our contribution is as follows.

\begin{itemize}
    \item We propose a differentially private scheme for federated learning with an adaptive gradient clipping and an adaptive noise scale reduction.
    \item We perform privacy loss analysis of our scheme using the notion of Rényi differential privacy (RDP) and prove that it satisfies the differential privacy.
    \item We conduct experimental comparisons on real data and show that our scheme is significantly better than previous methods.
\end{itemize}

The rest of this paper is organized as follows: Section II is the related work. Section III describes the preliminaries. In Section IV, we present our approach in detail, and conduct the privacy analysis. Section V shows the experimental results and the related analysis. Conclusion of this paper in Section VI. A summary of basic concepts and notations is provided in TABLE. I.
\begin{table}[t]
\caption{Summary of Main Notations}
\begin{center}
\begin{tabular}{ll}
\\ \hline \\
$K$ & \text {Total number of all clients } \\
$k$ & \text {The index of the $k$-th clients } \\
$T$ & \text {Total number of aggregation times } \\
$L$ & \text {Lot size for local training once } \\
$t$ & \text {The index of the $t$-th aggregation } \\
$D$ & \text {The database held by all the clients } \\
$D^k$ & \text {Loacl database of the $k$-th client } \\
$F^k(w)$ & \text{Local loss funtcion from the $k$-th clinet}\\
$w$ & \text {Global model} \\
$w^k$ & \text {Loacl model of the $k$ -th client } \\
$w^k_t$ & \text {Loacl model of the $k$ -th client in $t$-th aggregation} \\
$\eta_t$ & \text{Learning rate of the $t$-th aggregation }  \\
$p^k$ & \text {Model weights of the $k$-th client } \\
$\sigma^k$ & \text {Noise scale of the $k$-th client} \\
$J(w)$ & \text {Verification loss of the global model} \\
$L^k_t$ & \text {Sampling lot of the $k$-th client in $t$-th aggregation} \\
$C^k_t$ & \text {Clipping thresholds of the $k$-th client in $t$-th aggregation} \\
$\epsilon^k_t$ & \text {Privacy loss of the $k$-th client in $t$-th aggregation} \\
$\epsilon$ & \text {Global privacy budget} \\
$\beta$ & \text {Decay factor of the adaptive noise scale reduction method} \\
$\alpha$ & \text {Clipping factor of the adaptive gradient clipping method} \\
\\ \hline \\
\end{tabular}
\end{center}
\end{table}

\section{Related Work}
In the past research, two definitions of differential privacy are applied in federated learning: client-level differential privacy (CL-DP) and sample-level differential privacy (SL-DP).
\begin{itemize}
    \item \textbf{SL-DP:} In SL-DP setting, each client holds a set of samples, and adding one sample to or removing one sample from a client does not matter the output obviously\cite{MartnAbadi2016DeepLW,XiWu2016BoltonDP}. To protect the sample participation information under the client, The work in  \cite{StaceyTruex2018AHA,QingLiao2020DPFLAN,QinqingZheng2021FederatedP,MaxenceNoble2022DifferentiallyPF} proposed per-sample gradient clipping to obtain the sensitivity and add Gaussian noise to the local stochastic gradient. Nicolas Papernot et al.\cite{NicolasPapernot2016SemisupervisedKT} proposed a PATE approach based on knowledge aggregation by training a set of "teacher" models with disjoint data to answer "student" models whose attributes may be made public.
    \item \textbf{CL-DP:} In CL-DP setting, each client provides one data sample, and adding or removing a client does not matter the output obviously\cite{HBrendanMcMahan2017LearningDP}. The most common approach is to add noise after the central party has collected and aggregated the models of each client \cite{RobinCGeyer2017DifferentiallyPF}. Furthermore, The work in \cite{KangWei2021UserLevelPF,KangWei2019FederatedLW} assume that the central party is not credible, sets clipping bounds on each client's model parameters and directly applies perturbations to the local models.
\end{itemize}

There has been work on adaptive noise of differentially privacy under centralized deep learning in recent years. The work in\cite{KoenLennartvanderVeen2018ThreeTF} proposed a gradient-aware clipping scheme which adaptively clipped the different layers of gradient, but no privacy analysis is given. Zhiying Xu et al. \cite{ZhiyingXu2019AnAA} combined with the root mean square prop (RMSProp) gradient descent technique to adaptive add noise for coordinates of the gradient based on the average square of historical gradient. Lei Yu et al. \cite{LeiYu2019DifferentiallyPM} applied the pre-set equation to adaptive noise scale reduction as the number of rounds increases. Jaewoo Lee et al. \cite{JaewooLee2018ConcentratedDP} assigned different privacy budgets to each training round to mitigate the effect of noise on the gradient. The work of \cite{SashankJReddi2019AdaCliPAC} proposed a method for coordinate-wise adaptive clipping for DP-SGD.
The adaptive noise described above is in centralized deep learning. Although adaptive clipping to the median update norm is used in federated learning\cite{GalenAndrew2019DifferentiallyPL}, it is based on the idea of CL-DP, which performs noise adding after centralized party aggregation.

As a result, we are the first paper to combine and apply adaptive gradient clipping and adaptive noise scale reduction in SL-DP of federated learning.

\section{PRELIMINARIES}

\subsection{Federated learning}

Considering a general federated learning (FL) system with one server and \emph{K} clients, as shown in Fig. 2. Let $D^k$ denote the local database held by the $k$-th client, where  $k \in \{1,2,...,\emph{K}\}$. The goal of the server is to learn a model over data from the \emph{K}  associated clients, and to minimize the loss function of this model. Formally, the server adds the weights received from the \emph{K} clients as

\begin{equation}
w = \sum\limits_{i=1}^K p^k w^k
\end{equation}
where $ {w}^k$ is the parameter vector trained at the $k$-th client, $w$ is the parameter vector after aggregating at the server, $K$ is the number of clients, ${p^k}$ = $\frac{|D^k|}{|D|}$ $\geq$ 0 \, with \, $\sum\limits_{i=1}^K$ ${p^k}$ = 1, and $|D|$ $=$ $\sum\limits_{i=1}^K$ $|D^k|$ is the total size of all data samples. Such an optimization problem can be formulated as

\begin{equation}
w^* = {\rm {arg \, \mathop{min}_w}} \sum\limits_{i=1}^K p^k F^k (w,D^k)
\end{equation}
where $F^k(\cdot)$ is the local loss function of the $k$-th client. In the FL process, $K$ clients collaborate to learn the ML model with the help of the cloud server and finally optimize the equation (2) to converge to the solution of the global optimal learning model.

The FL can be categorized into two typical types by the scale of federation:
\begin{itemize}
    \item \textbf{Cross-device FL:} In cross-device FL, the number of parties is relatively large and each party has a relatively small amount of data as well as computational power. The parties are usually mobile devices. Google Keyboard is an example of cross-device FL\cite{HubertEichner2018FederatedLF}.
    \item \textbf{Cross-silo FL:} In cross-silo FL, for example in Fig.2, the participants are usually individual organizations or data centers, usually with relatively few participants, and each participant has a larger amount of data and computational power. For example, different hospitals can use cross-silo FL to train CNNs for chest film classification while exposing their chest film images locally\cite{TheodorosSalonidis2019DifferentialPF}. In this paper, we will apply SL-DP in cross-silo FL.
\end{itemize}

\begin{figure}[htbp]
	\begin{center}
		\includegraphics[width=\linewidth]{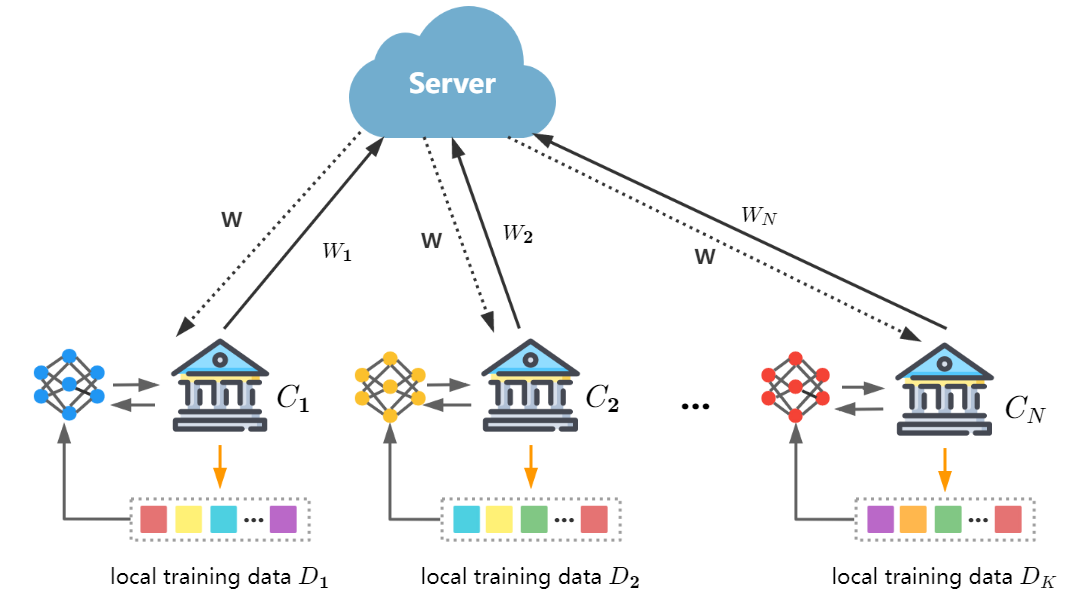}
		\caption{Federated Learning Model}
		\label{gra1}
	\end{center}
\end{figure}

\subsection{Differential privacy}

Differential privacy is a rigorous mathematical framework that formally defines the privacy loss of data analysis algorithms. Informally, it requires that any changes to a single data point in the training dataset can not cause statistically significant changes in the algorithm's output.

\begin{definition}
	(Differential Privacy\cite{dwork2014algorithmic}). The randomized mechanism A provides ($\epsilon$,  $\delta$) differential privacy, if for any two neighboring database D and D' that differ in only a single entry, $\forall$S $\subseteq$ Range(A),
\end{definition}
\begin{equation}
{\rm Pr}(A(D) \in S) < e^{\epsilon} {\rm Pr}(A(D') \in S) + \delta
\end{equation}

If $\delta$ = 0, $A$ is said to be $\epsilon$-differential privacy. In the rest of this paper,  we write ($\epsilon$, $\delta$)-DP for short.
Generally, 1 / $|D|$ is selected for $\delta$, and the default $\delta$ is $10^{- 5}$ in this paper.

\begin{definition}
	($R\acute{e}nyi$ Divergence\cite{IlyaMironov2017RnyiDP}).Given two probability distributions P and Q,The $R\acute{e}nyi$  divergence between P and Q with order $\alpha$ > 1 is defined by
	\begin{equation}
	\mathrm{D}_{\alpha}(P||Q) = \frac{1}{\alpha-1} \log \mathbb{E}_{x \sim Q}
	\left(\frac{P(x)}{Q(x)}\right)^{\alpha}
	\end{equation}
\end{definition}

\begin{definition}
($R\acute{e}nyi$ differential privacy (RDP)\cite{IlyaMironov2017RnyiDP}). A randomized mechanism $f : D \rightarrow \mathbb{R}$ is said to be $(\alpha,\epsilon)-R\acute{e}nyi$ differential private if for any adjacent datasets $D, D^{'}$, it holds that
	\begin{equation}
    \mathrm{D}_{\alpha}(f(D)||f(D^{'}))\leq \epsilon
	\end{equation}
\end{definition}

\subsection{Threat model}
We assume that the server is honest-but-curious, meaning that it will follow the rules of the federation but it is curious about the local data of each client. Although each client's individual dataset is kept locally in FL, the model parameter must be shared with the server, which may reveal the clients' private information by model-inversion attacks. This is reflected that 
the server want recover the training datasets\cite{ZhiboWang2018BeyondIC} or infer private features\cite{LucaMelis2022ExploitingUF} based on the local uploaded parameters. Further, it want deduce whether a sample belongs to the client's training dataset from the difference in the output\cite{RezaShokri2016MembershipIA}.

In addition, each client can also be seen honest-but-curious, and they can perform the above privacy attack on other client data after getting the model parameters broadcasted by the server.


\section{Our Approcah}
In this chapter, we will introduce the motivation and methods of adaptive gradient clipping and adaptive noise scale reduction, as well as describe the algorithm of Adap DP-FL after combining the two methods, and finally perform the privacy loss analysis of Adap DP-FL.
\subsection{Adaptive gradient clipping }
Per-sample gradient clipping is essential in DP-SGD\cite{MartnAbadi2016DeepLW}, and similarly in SL-DP with federated learning. In Per-sample gradient clipping, if the gradient $l_2$-norm of the sample involved in the training is more than C, each gradient entry of that sample is scaled down by a factor of C divided by the gradient $l_2$-norm of that sample. By per-sample gradient clipping, we can get the sensitivity of the gradient aggregate with respect to the addition or removal of each sample, thus adding Gaussian noise to the aggregated gradient and thus achieving central differential privacy guarantee.

The challenge for differential private training is selecting a suitable clipping threshold C\cite{MartnAbadi2016DeepLW}. If the clipping threshold C is too small, the average clipped gradient may point in a very different direction than the true gradient. If the clipping threshold C is too large, we have to add more noise to the gradient (parameter) because the variance of Gaussian noise is $\sigma * C$. Thus setting C either too high or too low can adversely affect the utility of the learned model. In addition, Fig. 1 shows that in federated learning, the gradient $l_2$-norm vary with different client ends and training rounds.



Based on the motivation and observed stated above, we propose an adaptive gradient clipping scheme for federated learning. For the $k$-th client, the method uses the $differentially$ $private$ $mean$ $gradient$ $l_2$-$norm$ $of$ $the$ $previous$ $round$ $times$ $\alpha$ as the clipping threshold for the current round, which $\alpha$ is a constant clipping factor. The clipping threshold $C^k_t$ for round $t$ and client $k$ is defined as follows, when $g^k_{t-1}$ is the individual gradient of the previous round and $\sigma^k_{t-1}$ is the individual noise scale of the previous round.
\begin{equation}
C_{t}^{k}=\alpha * ||\frac{\sum_{i \in L_{t}^k} \operatorname{clip}\left(\left\|g_{t-1}^{k}(x_i))\right\|_{2}\right)+\mathcal{N}\left(0, (S_{t-1}^k)^2 \right)}{L}||_2
\end{equation}
where
\begin{equation}
\operatorname{clip}\left(\|g_{t}^k(x_i)\|_{2}\right)=\|g_{t}^k(x_i)\|_{2} / \max \left(1, \frac{\|g_{t}^k(x_i)\|_{2}}{C_{t-1}^{k}}\right)
\end{equation}

\begin{equation}
S_{t-1}^k={C_{t-1}^{k}}*{\sigma_{t-1}^{k}}
\end{equation}

From the above, we can see that in order to ensure that the clipping satisfies differential privacy, we use the noise-added gradient as the clipping threshold $C_{t}^k$. Same reason, we use a similar adaptive procedure as described above to select $C_{t-1}^k$, ensuring that the clipping threshold $C$ in each round satisfies the post-processing nature of differential privacy. In particular, we initialize $C_0^k$ by training on random noise for one round and extracting the mean $l_2$-norm.

\subsection{Adaptive noise scale reduction}
In DP-SGD, Gaussian noise with variance $\sigma * C$ is added to each round, where $\sigma$ is the noise scale which is directly related to the privacy loss. Given a total privacy budget, if $\sigma$ is large, the privacy loss caused by each round will be small, meaning that more rounds can be performed, but the model will be affected by more noise. Whereas, if $\sigma$ is small, the privacy loss caused by each round is large, and although the model is less affected by noise, it will cause the problem of not enough rounds.

From the Fig. 1, we can find that the gradient $l_2$-norm will gradually decrease with rounds. So, we argue that noise has less of an effect on the model's gradient in the early stages of federated training, and that adding random perturbations to the random gradient descent in the pre-training period allows the gradient to quickly escape the saddle point \cite {RongGe2015EscapingFS}. During the training process, the effect of noise on the model gradually increases as the gradient becomes smaller and smaller. So, In the late training period, too much noise prevents the model from converging, and a small noise scale is required to bring the model close to optimization.
A similar strategy is used for the learning rate of DNNs. Instead of using a constant learning rate in all epochs to achieve better accuracy, it is usually recommended to reduce the learning rate as training progresses in later epochs\cite{JohnCDuchi2010AdaptiveSM,NingQian1999OnTM}.

In general, our adaptive noise sacle reduction follows the idea that, as the central model accuracy converges, it is expected to have less noise on the gradient of each client, which allows the learning process to get closer to the local optimal spot and achieve better accuracy. Our method adjusts the noise scale of each client dynamically based on the variation of the verification loss on the central side. Let $t$ denote the round of each client round, $\sigma_t$ denote the noise scale of $t$ training for each client, and $J(w_t)$ denote the verification loss of global model of $t$-th round. Initial, $J(w_0)=J(w_1)=J(w_2)=\infty$. After that, until the next validation round $t+1$, the updated noise scale $\sigma_{t+1}$ is applied to the training.
\begin{equation}
\sigma_{t+1}=\left\{
\begin{array}{rcl}
\beta * \sigma_t  & {J(w_{t-3})>J(w_{t-2})>J(w_{t-1})>J(w_{t})}\\
\sigma_t & else \\
\end{array} \right.
\end{equation}

When the server detects that the loss value of the validation set has dropped for the third time in a row, the noise scale of all clients decays $\sigma_{t+1}=\beta*\sigma_t$, where $\beta\ (0<\beta<1)$ is the decay rate, a schedule hyperparameter.
We note that the central validation loss may not decrease monotonically as the training progresses, and its fluctuations may cause unnecessary reduction of noise scale and thus waste on the privacy budget. This motivates us to use three consecutive decreases in the loss value as a criterion for the noise scale decrease.

\subsection{Adap DP-FL}
We combine the two methods mentioned above and apply them to DP-FL to create our Adap DP-FL, the training process of which is shown in Algorithm 1, and consists of the following steps:

\begin{algorithm}
	\renewcommand{\algorithmicrequire}{\textbf{Input:}}
	\renewcommand{\algorithmicensure}{\textbf{Output:}}
	\caption{Adap DP-FL}
	\label{alg:1}
	\begin{algorithmic}[1]
	\REQUIRE {gradient clipping factor $\alpha$, noise reduction factor $\beta\ (0<\beta<1)$ , group size $L$ ,learning rate $\eta_t$, inital noise scale $\sigma_t$, global privacy budget $({\epsilon},\delta)$}
		\STATE Initialize: $t=0$ and $w_0$
		\WHILE {$t<T$}
		\STATE {Broadcast: $w_t$ and $\sigma_t$ to all clients}
		\STATE \textbf {Local training process:}
		\FOR{$\forall k \in K$}
		\STATE \textbf{Compute local gradient:}
		\STATE Take a random sample $L_t^k$ with sampling probability $L/|D^k|$
		\STATE for each $i\in L_t^k$,compute $g_t^k(x_i)\leftarrow \nabla \zeta{(w_t^k,x_i)}  $
		\STATE \textbf{Adaptive gradient clipping:}
		\STATE Each client computes clipping threshold ${C_t^k}$ by equation (6), (7) and (8)
		\STATE $\overline{g}_t^k(x_i) \leftarrow g_t^k{(x_i)} / max(1,\frac{||g_t^k{(x_i)}_2||}{C_t^k}) $
		\STATE \textbf{Add Gaussian noise:}
		\STATE $\widetilde{g}_t^k \leftarrow \frac{1}{L} (\sum\limits\nolimits_i \overline{g}_t^k(x_i)+\mathcal{N}(0,{(\sigma_t)}^2 {(C_t^k)}^2)) $
		\STATE \textbf{gradient descent:}
		\STATE $w_{t+1}^k=w_t^k - \eta_t \widetilde{g}_t^k $
		\STATE \textbf {Compute privacy loss $\epsilon_t^k$ by equation (10)}
		\IF{$\epsilon_t^k>\epsilon$}
		\STATE{break}
		\ENDIF
		\STATE{upload model parameters to the server}
		\ENDFOR
		\STATE \textbf {Model aggregating process:}
		\STATE $w_{t+1} = \sum\limits\nolimits_{k\in K} p^k w_{t+1}^k$
		\IF {$ J(w_{t-2})>J(w_{t-1})>J(w_{t})>J(w_{t+1})$}
		\STATE $\sigma_{t+1}$ = $\beta$$\sigma_t$
		\ENDIF
		\STATE $t=t+1$
\ENDWHILE
		\STATE \textbf{Output} $w_t$ and privacy loss $\epsilon_t^k$ of the $k$-th client
	\end{algorithmic}
\end{algorithm}

\begin{itemize}
	\item \textbf{Initialization}:
	The server initializes and broadcasts global model $w^0$, initial noise scale $\sigma_0$, clipping factor $\alpha$, noise reduction scale $\beta$ and global privacy budget $(\epsilon,\delta)$ to all clients.
	\item \textbf{Step 1: $Local\ gradient\ computing$}: At each client, a lot $L^k_t$ is a sample from the local training dataset $D^k$. Then we compute the gradient with per-example gradient algorithm.
	\item \textbf{Step 2: $Adaptive\ gradient\ clipping$}: Calculate the clipping threshold $C_t^k$ for the $k$-th client in the $t$-th round, and clip per-sample gradient by $C_t^k$. 
	\item \textbf{Step 3: $Noise\ adding$}: Each client adds artificial noise according to the differential privacy rule of the Gaussian mechanism, the standard deviation of the Gaussian noise of the \emph{k}-th client in the \emph{t}-th round is $C_t^k*\sigma_t$.
	\item \textbf{Step 4: $Gradient\ descent$}:Each client performs gradient descent with learning rate to obtain the model $w_{t+1}^k$.
	\item \textbf{Step 5: $Privacy\ loss\ calculation$}: For each client, the privacy loss is calculated according to the noise scale of this round by using Theorem 1.
	\item \textbf{Step 6: $Model\ uploading$}: Each client uploads the model to the server provided that the privacy loss meets the privacy budget.
	\item \textbf{Step 7: $Model\ aggregation$}: The server weighted the models uploaded by each client and averaged them.
	\item \textbf{Step 8: $Adaptive\ noise\ scale\ reduction$}: The server executes Equation 8 for adaptive noise scale reduction to get the $\sigma_{t+1}$ for the next round to all clients.
	\item \textbf{Step 9: $Model\ broadcasting$}: The server broadcasts global model and noise scale to each client.
\end{itemize}

\subsection{Privacy Loss Analysis}
Each client adds Gaussian noise satisfying differential privacy to the locally trained model can defend against the attacks mentioned in the threat model above. As a result, we only assess each client's privacy loss from the perspective of a single client. Second, because each client's locally trained pair of data samples has the same sampling rate and noise factor in each round, each client has the same privacy loss in each round. Following that, we examine the privacy budget for the $k$-th client (the privacy loss analysis for the other clients is the same as it is).
\begin{theorem}(DP Privacy Loss of Algorithm 1). After T rounds, the $k$-th client DP Privacy budget of Algorithm 1 satisfies:
\begin{equation}
\begin{split}
    (\epsilon^k_T,\delta)=(\sum_{t=0}^T \frac{1}{\alpha-1}\sum_{i=0}^{\alpha}\left(\begin{array}{l}
\alpha \\ i
\end{array}\right)(1-q)^{\alpha-i} q^{i}\\ \exp \left(\frac{i^{2}-i}{2 \sigma_t^{2}}\right)
+ \frac{\log 1/\delta}{\alpha-1},\delta)
\end{split}
\end{equation}
where $q=\frac{L}{|D^k|}$ , $\sigma_t$ is noise scale of $t-th$ iters and any integer $\alpha \geq 2$ .
\end{theorem}
$proof.$ We perform privacy loss calculation based on RDP\cite{IlyaMironov2017RnyiDP}. We first use the sampling Gaussian theorem of RDP to calculate the privacy cost of each round, then we perform the advanced combination of RDP to combine the privacy cost of multiple rounds, and finally we convert the obtained to RDP privacy to DP.
\begin{definition}
(Sampled Gaussian Mechanism (SGM)\cite{IlyaMironov2019RnyiDP}). Let $f$ be a function mapping subsets of $S$ to $\mathbb{R}^d$. We define the Sampled Gaussian Mechanism (SGM) parameterized with the sampling rate $0 < q \leq 1$ and the  $\sigma > 0$ as
\begin{equation}
\begin{aligned}
	S G_{q, \sigma}(S) \triangleq & f(\{x: x \in S \text { is sampled with probability } q\}) \\
	&+\mathcal{N}\left(0, \sigma^{2} \mathbb{I}^{d}\right)
	\end{aligned}
\end{equation}
in DP-SGD, $f$ is the clipped gradient evaluation in sampled data points $f(\{x_i\}_{i\in B}) = \sum_{i\in B} \overline{g}_t(x_i)$. If $ \overline{g}_t$ is obtained by clipping $g_t$ with a gradient norm bound $C$, then the sensitivity of $f$ is equal to $C$.
\end{definition}

\begin{definition}
(RDP privacy budget of SGM\cite{IlyaMironov2019RnyiDP}). Let $SG_{q,\sigma}$, be the Sampled Gaussian Mechanism for some function $f$. If $f$ has sensitivity 1, $SG_{q,\sigma}$ satisfies $(\alpha,\epsilon)$-RDP whenever
\begin{equation}
\epsilon \leq \frac{1}{\alpha-1} \log max(A_{\alpha}(q,\sigma),B_{\alpha}(q,\sigma))
\end{equation}
where
\begin{equation}
\left\{\begin{array}{l}
A_{\alpha}(q, \sigma) \triangleq \mathbb{E}_{z \sim \mu_{0}}\left[\left(\mu(z) / \mu_{0}(z)\right)^{\alpha}\right] \\
B_{\alpha}(q, \sigma) \triangleq \mathbb{E}_{z \sim \mu}\left[\left(\mu_{0}(z) / \mu(z)\right)^{\alpha}\right]
\end{array}\right.
\end{equation}
with $\mu_{0} \triangleq \mathcal{N}\left(0, \sigma^{2}\right), \mu_{1} \triangleq \mathcal{N}\left(1, \sigma^{2}\right) \mbox { and } \mu \triangleq(1-q) \mu_{0}+q \mu_{1}$

Further, it holds $\forall(q,\sigma)\in(0,1], \mathbb{R}^{+*},A_{\alpha}(q,\sigma) \geq B_{\alpha}(q, \sigma) $. Thus, $ S G_{q, \sigma}  satisfies  \left(\alpha, \frac{1}{\alpha-1} \log \left(A_{\alpha}(q, \sigma)\right)\right)$-RDP .

Finally, \cite{IlyaMironov2017RnyiDP} describes a procedure to compute $A_{\alpha}(q,\sigma)$ depending on integer $\alpha$.
\begin{equation}
A_{\alpha}=\sum_{k=0}^{\alpha}\left(\begin{array}{l}
\alpha \\ k
\end{array}\right)(1-q)^{\alpha-k} q^{k} \exp \left(\frac{k^{2}-k}{2 \sigma^{2}}\right)
\end{equation}
\end{definition}

\begin{definition}
(Composition of RDP\cite{IlyaMironov2017RnyiDP}). For two randomized mechanisms $f, g$ such that $f$ is $(\alpha,\epsilon_1)$-RDP and $g$ is $(\alpha,\epsilon_2)$-RDP the composition of $f$ and $g$ which is defined as $(X, Y )$(a sequence of results), where $ X \sim f $ and $Y \sim g$, satisfies $(\alpha,\epsilon_1+\epsilon_2)-RDP$
\end{definition}

From Definition 4, Definition 5 and Definition 6, the following Lemma 1 is obtained.
\begin{lemma}. Given the sampling rate $q=\frac{L}{N}$ for each round of the local dataset and $\sigma_t$ as the noise factor for round t, the total RDP privacy loss of the kth client for round T loss for any integer $\alpha \geq 2$ is:
\begin{equation}
    \epsilon^{'}(\alpha)_T^k=\sum_{t=0}^T \frac{1}{\alpha-1}\sum_{i=0}^{\alpha}\left(\begin{array}{l}
\alpha \\ i
\end{array}\right)(1-q)^{\alpha-i} q^{i} \exp \left(\frac{i^{2}-i}{2 \sigma_t^{2}}\right)
\end{equation}
\end{lemma}

\begin{definition}
(Translation from RDP to DP\cite{IlyaMironov2017RnyiDP}). if a randomized mechanism $f : D \rightarrow \mathbb{R}$  satisfies $(\alpha,\epsilon)$-RDP ,then it satisfies$(\epsilon +\frac{\log 1/\delta}{\alpha-1},\delta)$-DP where $0<\delta<1$
\end{definition}

By Lemma 1 and Definition 7, Theorem 1 is proved. In practice, given $\sigma$, $\delta$ and $q$ at each round, we select $\alpha$ from $\left\{2,3,...,64\right\}$ and determine the smallest  $\epsilon_{*}$ in Theorem 1. The privacy loss is the pair $(\epsilon_{*},\delta)$.

\section{EXPERIMENT}
We conducted a series of comparative experiments on Mnist datasets and FashionMnist datasets. As a benchmark for comparison, we use the algorithm of \cite{QinqingZheng2021FederatedP}, which employs a constant clipping threshold and noise scale for all clients. Experiments show that our Adap DP-FL method performs better than previous methods.

\begin{figure*}[htbp]
	\centering
	\subfigure[avg test loss result in Mnist datasets]{
	\includegraphics[width=0.20\textwidth,height=3.0cm]{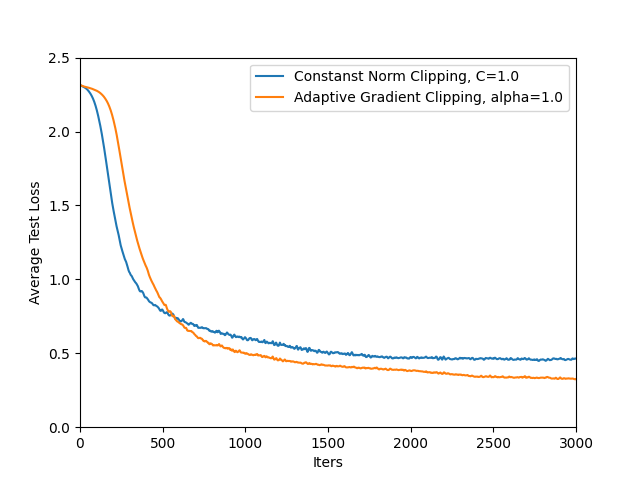}
	}
	\quad
	\subfigure[avg test acc result in Mnist datasets]{
	\includegraphics[width=0.20\textwidth,height=3.0cm]{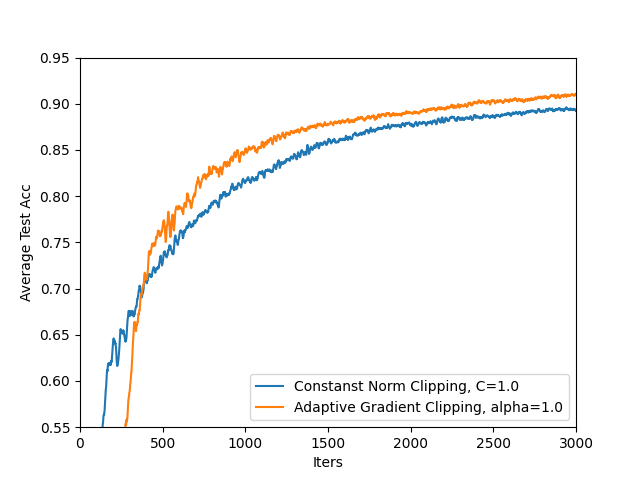}
	}
	\quad
	\subfigure[avg test loss result in FashionMnist datasets]{
	\includegraphics[width=0.20\textwidth,height=3.0cm]{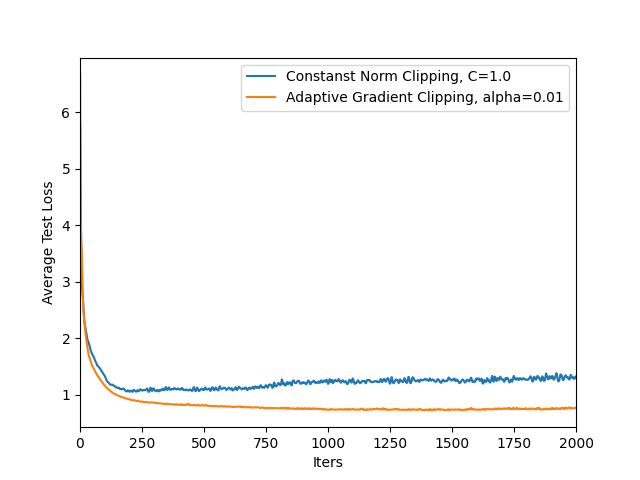}
	}
	\quad
	\subfigure[avg test acc result in FashionMnist datasets]{
	\includegraphics[width=0.20\textwidth,height=3.0cm]{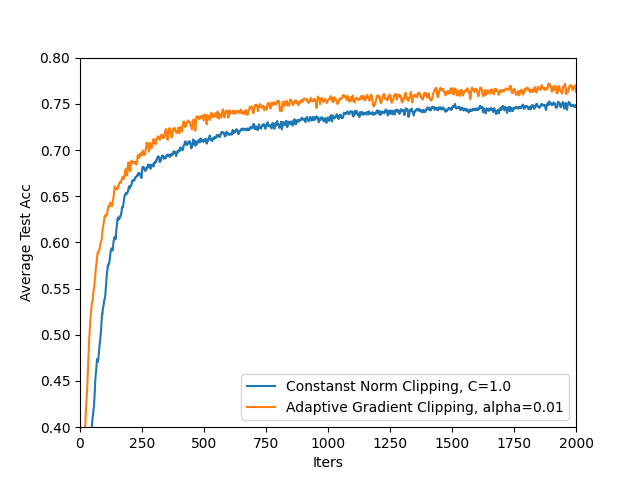}
	}
	\caption{Performance for adptivate gradient clipping method on Mnist dataset and FashionMnist dataset.}

\end{figure*}

\begin{figure*}[htbp]
	\centering
	\subfigure[avg test loss result in Mnist datasets]{
	\includegraphics[width=0.20\textwidth,height=3.0cm]{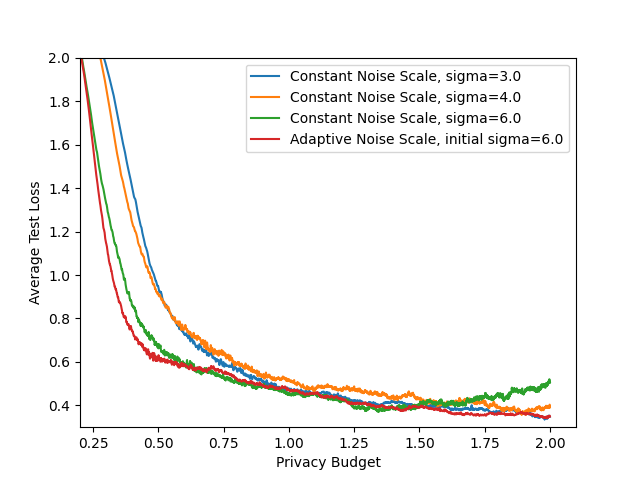}
	}
	\quad
	\subfigure[avg test acc result in Mnist datasets]{
	\includegraphics[width=0.20\textwidth,height=3.0cm]{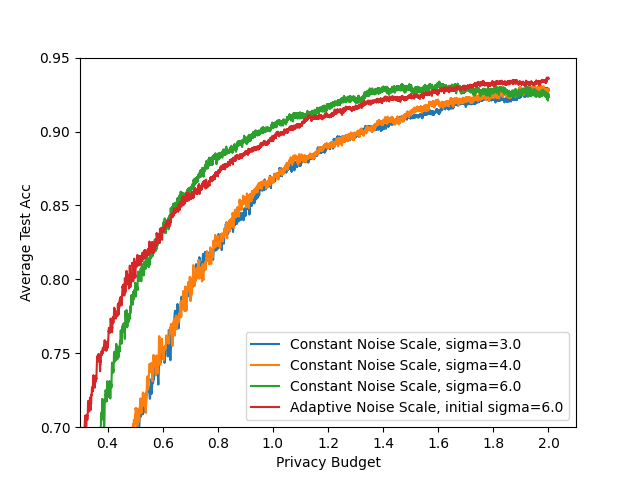}
	}
	\quad
	\subfigure[avg test loss result in FashionMnist datasets]{
	\includegraphics[width=0.20\textwidth,height=3.0cm]{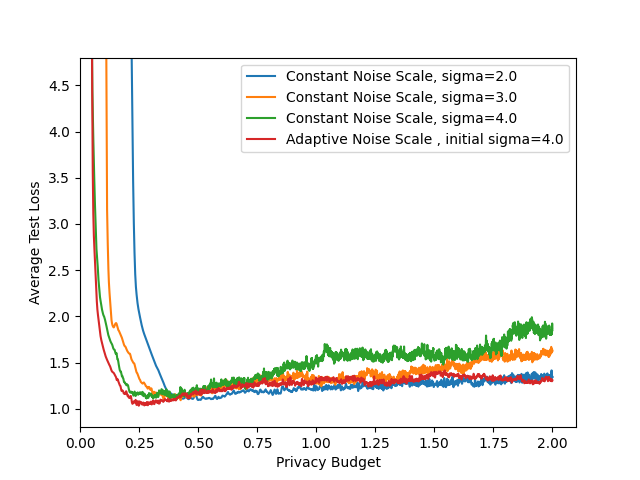}
	}
	\quad
	\subfigure[avg test acc result in FashionMnist datasets]{
	\includegraphics[width=0.20\textwidth,height=3.0cm]{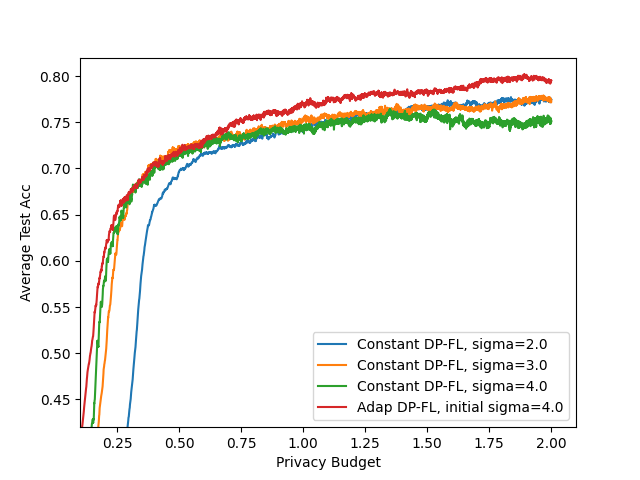}
	}
	\caption{Performance for adptivate noise scale reduction method on Mnist dataset and FashionMnist dataset.}
\end{figure*}

\begin{figure*}[htbp]
	\centering
	\subfigure[avg test loss result in Mnist dataset]{
	\includegraphics[width=0.2\textwidth,height=3.0cm]{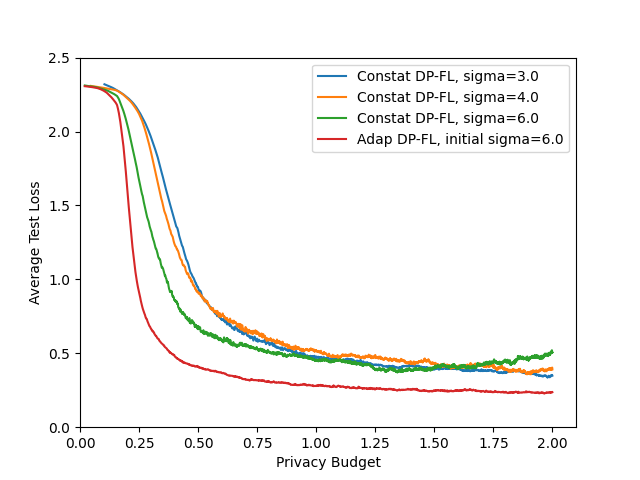}
	}
	\quad
	\subfigure[avg test acc result in Mnist dataset]{
	\includegraphics[width=0.2\textwidth,height=3.0cm]{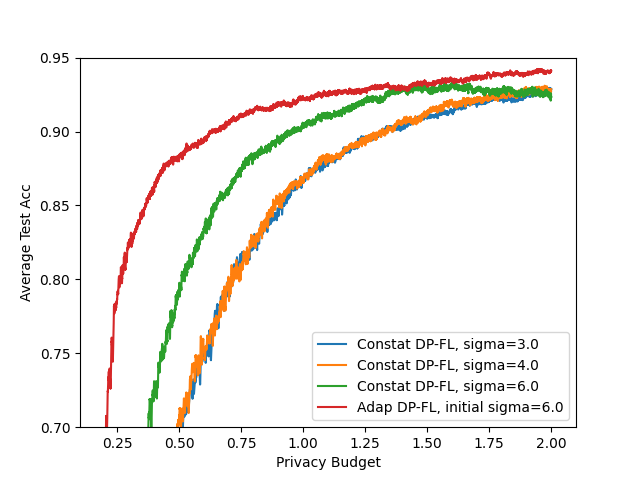}
	}
	\quad
	\subfigure[avg test loss result in FashionMnist dataset]{
	\includegraphics[width=0.2\textwidth,height=3.0cm]{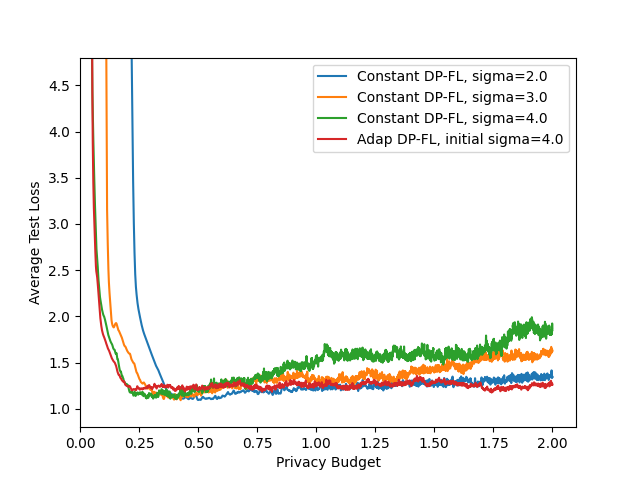}
	}
	\quad
	\subfigure[avg test acc result in FashionMnist dataset]{
	\includegraphics[width=0.2\textwidth,height=3.0cm]{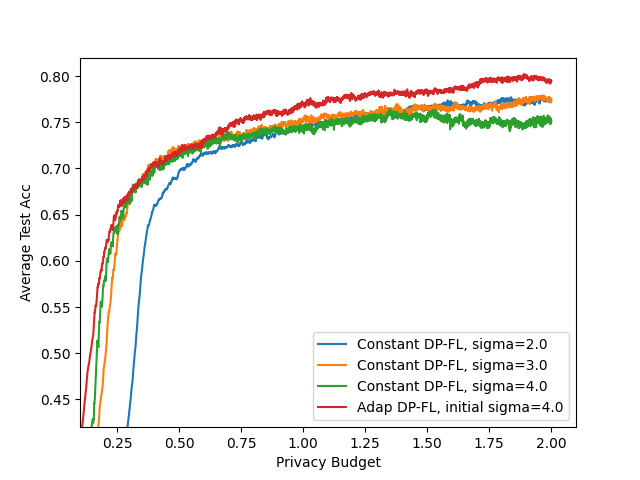}
	}
	\caption{Performance for Adap DP-FL method on Mnist dataset and FashionMnist dataset. The clipping factor $\alpha$=1.0 and noise reduction factor $\beta$=0.9999 for Adap DP-FL in Mnist datasets. The clipping factor $\alpha$=0.01 and noise reduction factor $\beta$=0.9998 for Adap DP-FL in FashionMnist datasets.}
\end{figure*}

\subsection{initialize}
\textbf{Dataset and partition settings.} 
Mnist\cite{LiDeng2012TheMD} and FashionMnist\cite{AlexKrizhevsky2009LearningML} are the standard dataset for handwritten digit recognition and clothing classification, respectively. Both of them consist of 60,000 training samples and 10,000 test samples, and each sample is a $28\times28$ grayscale image.
We use a similar setup to MHR et al \cite{HBrendanMcMahan2017LearningDP} to perform Non-IID partitioning on the data from 10 clients. We sorted the training data according to numerical labels and partitioned it equally into 400 fragments. Each of the 10 clients was assigned 40 random data fragments, so that most clients had samples with three or four less-than-identical labels, and individual clients had samples with two or five labels.

\textbf{Network model.} 
We use a shallow neural network which consists of two convolutional layers adaptived by rectified linear units (ReLU), and two fully connected layers. The output channels of the two convolutional layers are 16 and 32, and the size of the convolutional kernels are $8 \times 2$ and $4 \times 2$. One fully connected layer has an input size of $32 \times 4 \times 4$ and an output size of 32, and the other fully connected layer has an input size of 32 and an output size of 10.

\textbf{Parameter settings.}
We default to one round per client local to participate in the federated average. We use the Adam optimizer with a base learning rate of 0.002 for the Mnist handwritten image recognition task and the Adam optimizer with a learning rate of 0.001 for the FashionMnist clothing classification task. We assigned each client a lot size of $L$=78.

\subsection{Performance}
\subsubsection{Adaptive gradient clipping}
Fig. 3 investigates the performance of adaptive gradient clipping method in DP-FL. It compares the average test loss and accuracy using adaptive gradient clipping method and constant gradient clipping method. We set the clipping factors $\alpha$ to 1.0 and 0.01 on the Mnist dataset and FashionMnist dataset, respectively. The constant gradient clipping method uses a clipping threshold of $C=1.0$. Both methods in the Mnist datasets and FashionMnist datasets use constant noise scale $\sigma=1.1$. For both datasets, the adaptive gradient clipping method outperforms significantly the constant factor clipping method.

\subsubsection{Adaptive noise scale reduction}
Fig. 4 investigates the performance of adaptive noise scale reduction method in DP-FL. It compares the average test loss and accuracy using adaptive noise scale reduction method and constant noise scale method.
In the Mnist dataset, As in Fig. 6, the noise scale decreases from 6 to 3.03 with privacy budget $\epsilon$=2.0 when the noise reduction factor $\beta$=0.9999.
So we compare adaptive noise scale reduction method and constant noise scale method with $\sigma$=3,4,6 in Mnist dataset. We set the constant clipping threshold $C$=1.0 for all method. From subfigure (a) and (b) in Fig. 4, we can find that if we use a large constant noise scale with $\sigma$=6, it can make the model perform well in the early stage of training, and when it reaches about 93\% accuracy (approaching convergence), it cannot further improve the accuracy or even the performance decreases in the subsequent training. In contrast, a small noise scale with $\sigma$=3 or $\sigma$=4 wastes more privacy budget in the early stage, but has the ability to approach convergence in the later stage of training. Our adaptive noise scale reduction method enables to save privacy budget to obtain higher model accuracy performance in the early stage, and further improve the model accuracy as the noise scale reduce in the later training stage. Especially, at privacy budget $\epsilon$=2, the average test accuracy by adaptive noise scale reduction method achieves 93.56\%, the average test accuracy by adaptive noise scale reduction method achieves 92.87\%, 92.86\% and 92.34\% when the $\sigma$=3.0, $\sigma$=4.0 and $\sigma$=6.0.

In the FashionMnist dataset, We set the initial noise sacle $\sigma$=4.0 and noise reduction factor $\beta$=0.9998. As in Fig. 6, the noise scale decreases from 4 to 2.11 when the privacy budget $\epsilon$=2.0.
So we compare adaptive noise scale reduction method and constant noise scale method with $\sigma$=2,3,4 in FashionMnist dataset. We set the constant clipping threshold $C=1.0$ for all method. From subfigure (c) and (d) in Fig. 4, we can find the similar performance as on the mnist dataset. Especially, at privacy budget $\epsilon$=2, the average test accuracy by adaptive noise scale reduction method achieves 78.48\%, the average test accuracy by adaptive noise scale reduction method achieves 77.28\%, 77.03\% and 75.14\% when the $\sigma$=2.0, $\sigma$=3.0 and $\sigma$=4.0.

\subsubsection{Adap DP-FL}
Finally, the Fig. 5 shows the performance of Adap DP-FL approach on Mnist datasets and FashionMnist datasets. Adap DP-FL combines the adaptive noise scale reduction method (experiment of Fig. 4) and the adaptive noise scale reduction method (experiment of Fig. 3). where the parameter settings are the same as those given in the above. We can see that Adap DP-FL performs better than the previous DP-FL method on both datasets. In particular, the Adap DP-FL approach was performed on the Mnist datasets and FashionMnist datasets, the average test accuracy achieves 94.15\% and 79.56\% when privacy budget $\epsilon$=2, respectively.

\begin{figure}[htbp]
	\centering
	\subfigure[ Mnist dataset] {
	\includegraphics[width=3.8cm,height=2.0cm]{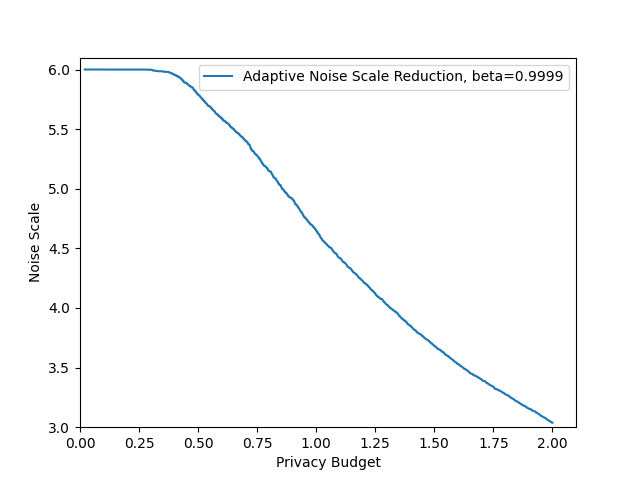}
	}
	\quad
	\subfigure[ FashionMnist dataset]{
	\includegraphics[width=3.8cm,height=2.0cm]{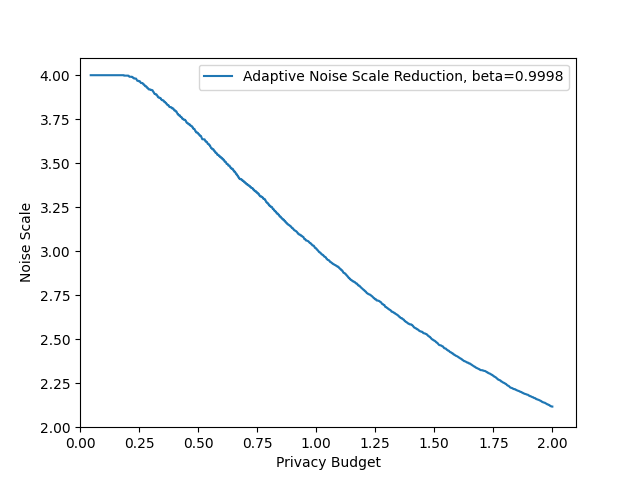}
	}
	\caption{ The variation of noise scale $\sigma$  for privacy budget in Adaptive noise scale reduction}
\end{figure}

\section{CONCLUSION}
In this paper, our main contributions are threefold. First, we propose the adaptive federated learning algorithm (Adap DP-FL), which has higher accuracy compared to previous algorithms. Second, we mathematically prove that our algorithm satisfies differential privacy through RDP. Third, we apply Adap DP-FL to train models on deep learning networks with real datasets and experimentally evaluate that Adap DP-FL has better performance relative to previous methods.
\bibliographystyle{IEEEtran}

\pdfoutput=1
\begin{thebibliography}{10}
\providecommand{\url}[1]{#1}
\csname url@samestyle\endcsname
\providecommand{\newblock}{\relax}
\providecommand{\bibinfo}[2]{#2}
\providecommand{\BIBentrySTDinterwordspacing}{\spaceskip=0pt\relax}
\providecommand{\BIBentryALTinterwordstretchfactor}{4}
\providecommand{\BIBentryALTinterwordspacing}{\spaceskip=\fontdimen2\font plus
\BIBentryALTinterwordstretchfactor\fontdimen3\font minus
  \fontdimen4\font\relax}
\providecommand{\BIBforeignlanguage}[2]{{%
\expandafter\ifx\csname l@#1\endcsname\relax
\typeout{** WARNING: IEEEtran.bst: No hyphenation pattern has been}%
\typeout{** loaded for the language `#1'. Using the pattern for}%
\typeout{** the default language instead.}%
\else
\language=\csname l@#1\endcsname
\fi
#2}}
\providecommand{\BIBdecl}{\relax}
\BIBdecl

\bibitem{mcmahan2017communication}
B.~McMahan, E.~Moore, D.~Ramage, S.~Hampson, and B.~A. y~Arcas,
  ``Communication-efficient learning of deep networks from decentralized
  data,'' pp. 1273--1282, 2017.

\bibitem{zhu2019deep}
L.~Zhu, Z.~Liu, and S.~Han, ``Deep leakage from gradients,'' \emph{Advances in
  neural information processing systems}, vol.~32, 2019.

\bibitem{song2017machine}
C.~Song, T.~Ristenpart, and V.~Shmatikov, ``Machine learning models that
  remember too much,'' pp. 587--601, 2017.

\bibitem{dwork2014algorithmic}
C.~Dwork, A.~Roth \emph{et~al.}, ``The algorithmic foundations of differential
  privacy,'' \emph{Foundations and Trends{\textregistered} in Theoretical
  Computer Science}, vol.~9, no. 3--4, pp. 211--407, 2014.


\bibitem{ZhiboWang2018BeyondIC}
Z.~Wang, M.~Song, Z.~Zhang, Y.~Song, Q.~Wang, and H.~Qi, ``Beyond inferring
  class representatives: User-level privacy leakage from federated learning,''
  \emph{international conference on computer communications}, 2018.


\bibitem{LucaMelis2022ExploitingUF}
L.~Melis, C.~Song, E.~D. Cristofaro, and V.~Shmatikov, ``Exploiting unintended
  feature leakage in collaborative learning,'' \emph{ieee symposium on security
  and privacy}, 2022.

\bibitem{MartnAbadi2016DeepLW}
M.~Abadi, A.~Chu, I.~Goodfellow, H.~B. McMahan, I.~Mironov, K.~Talwar, and
  L.~Zhang, ``Deep learning with differential privacy,'' \emph{computer and
  communications security}, 2016.

\bibitem{KangWei2019FederatedLW}
K.~Wei, J.~Li, M.~Ding, C.~Ma, H.~H. Yang, F.~Farhad, S.~Jin, T.~Q.~S. Quek,
  and H.~V. Poor, ``Federated learning with differential privacy: Algorithms
  and performance analysis,'' \emph{IEEE Transactions on Information Forensics
  and Security}, 2019.

\bibitem{HBrendanMcMahan2017LearningDP}
H.~B. McMahan, D.~Ramage, K.~Talwar, and L.~Zhang, ``Learning differentially
  private recurrent language models,'' \emph{Learning}, 2017.

\bibitem{NicolasPapernot2016SemisupervisedKT}
N.~Papernot, M.~Abadi, {\'U}.~Erlingsson, I.~Goodfellow, and K.~Talwar,
  ``Semi-supervised knowledge transfer for deep learning from private training
  data,'' \emph{Learning}, 2016.

\bibitem{XiWu2016BoltonDP}
X.~Wu, F.~Li, A.~Kumar, K.~Chaudhuri, S.~Jha, and J.~F. Naughton, ``Bolt-on
  differential privacy for scalable stochastic gradient descent-based
  analytics,'' \emph{international conference on management of data}, 2016.

\bibitem{KangWei2021UserLevelPF}
K.~Wei, J.~Li, M.~Ding, C.~Ma, H.~Su, B.~Zhang, and H.~V. Poor, ``User-level
  privacy-preserving federated learning: Analysis and performance
  optimization,'' \emph{IEEE Transactions on Mobile Computing}, 2021.

\bibitem{QinqingZheng2021FederatedP}
Q.~Zheng, S.~Chen, Q.~Long, and W.~J. Su, ``Federated \$f\$-differential
  privacy,'' \emph{arXiv: Machine Learning}, 2021.

\bibitem{IlyaMironov2017RnyiDP}
I.~Mironov, ``R{\'e}nyi differential privacy,'' \emph{ieee computer security
  foundations symposium}, 2017.

\bibitem{IlyaMironov2019RnyiDP}
I.~Mironov, K.~Talwar, and L.~Zhang, ``R{\'e}nyi differential privacy of the
  sampled gaussian mechanism.'' \emph{arXiv: Learning}, 2019.

\bibitem{LiDeng2012TheMD}
L.~Deng, ``The mnist database of handwritten digit images for machine learning
  research,'' \emph{IEEE Signal Processing Magazine}, 2012.

\bibitem{AlexKrizhevsky2009LearningML}
A.~Krizhevsky, ``Learning multiple layers of features from tiny images,'' 2009.

\bibitem{KoenLennartvanderVeen2018ThreeTF}
K.~L. van~der Veen, R.~Seggers, P.~Bloem, and G.~Patrini, ``Three tools for
  practical differential privacy,'' \emph{arXiv: Machine Learning}, 2018.

\bibitem{ZhiyingXu2019AnAA}
Z.~Xu, S.~Shi, A.~X. Liu, J.~Zhao, and L.~Chen, ``An adaptive and fast
  convergent approach to differentially private deep learning,''
  \emph{international conference on computer communications}, 2019.

\bibitem{RobinCGeyer2017DifferentiallyPF}
R.~C. Geyer, T.~Klein, and M.~Nabi, ``Differentially private federated
  learning: A client level perspective,'' \emph{arXiv: Cryptography and
  Security}, 2017.


\bibitem{JohnCDuchi2010AdaptiveSM}
J.~C. Duchi, E.~Hazan, and Y.~Singer, ``Adaptive subgradient methods for online
  learning and stochastic optimization.'' \emph{Journal of Machine Learning
  Research}, 2010.

\bibitem{NingQian1999OnTM}
N.~Qian, ``On the momentum term in gradient descent learning algorithms,''
  \emph{Neural Networks}, 1999.

\bibitem{LeiYu2019DifferentiallyPM}
L.~Yu, L.~Liu, C.~Pu, M.~E. Gursoy, and S.~Truex, ``Differentially private
  model publishing for deep learning,'' \emph{ieee symposium on security and
  privacy}, 2019.

\bibitem{RongGe2015EscapingFS}
R.~Ge, F.~Huang, C.~Jin, and Y.~Yuan, ``Escaping from saddle points: Online
  stochastic gradient for tensor decomposition,'' \emph{Journal of Machine
  Learning Research}, 2015.

\bibitem{GalenAndrew2019DifferentiallyPL}
G.~Andrew, O.~Thakkar, H.~B. McMahan, and S.~Ramaswamy, ``Differentially
  private learning with adaptive clipping.'' \emph{arXiv: Learning}, 2019.

\bibitem{QingLiao2020DPFLAN}
Q.~Liao, S.~Qi, X.~Huang, Z.~L. Jiang, Y.~Ding, and X.~Wang, ``Dp-fl: a novel
  differentially private federated learning framework for the unbalanced
  data,'' \emph{World Wide Web}, 2020.


\bibitem{HubertEichner2018FederatedLF}
H.~Eichner, A.~S. Hard, S.~Augenstein, D.~Ramage, K.~Rao, C.~Kiddon,
  R.~Mathews, and F.~Beaufays, ``Federated learning for mobile keyboard
  prediction,'' \emph{arXiv: Computation and Language}, 2018.

\bibitem{TheodorosSalonidis2019DifferentialPF}
T.~Salonidis, A.~K. Das, I.~Sylla, A.~Gkoulalas-Divanis, O.~Choudhury, G.~Hsu,
  and Y.~Park, ``Differential privacy-enabled federated learning for sensitive
  health data,'' \emph{arXiv: Learning}, 2019.

\bibitem{JaewooLee2018ConcentratedDP}
J.~Lee and D.~Kifer, ``Concentrated differentially private gradient descent
  with adaptive per-iteration privacy budget,'' \emph{knowledge discovery and
  data mining}, 2018.

\bibitem{SashankJReddi2019AdaCliPAC}
S.~J. Reddi, S.~Kumar, A.~T. Suresh, F.~X. Yu, and V.~Pichapati, ``Adaclip:
  Adaptive clipping for private sgd.'' \emph{arXiv: Learning}, 2019.

\bibitem{StaceyTruex2018AHA}
S.~Truex, N.~Baracaldo, A.~Anwar, T.~Steinke, H.~Ludwig, R.~Zhang, and Y.~Zhou,
  ``A hybrid approach to privacy-preserving federated learning,''
  \emph{computer and communications security}, 2018.

\bibitem{StaceyTruex2020LDPFedFL}
S.~Truex, L.~Liu, K.-H. Chow, M.~E. Gursoy, and W.~Wei, ``Ldp-fed: federated
  learning with local differential privacy,'' \emph{european conference on
  computer systems}, 2020.

\bibitem{RezaShokri2016MembershipIA}
R.~Shokri, M.~Stronati, C.~Song, and V.~Shmatikov, ``Membership inference
  attacks against machine learning models,'' \emph{arXiv: Cryptography and
  Security}, 2016.

\bibitem{MaxenceNoble2022DifferentiallyPF}
M.~Noble, A.~Bellet, and A.~Dieuleveut, ``Differentially private federated
  learning on heterogeneous data,'' 2022.


\end{thebibliography}

\end{document}